# Proving the Potential of Skeleton Based Action Recognition to Automate the Analysis of Manual Processes

Berger, Marlin[1]; Cloppenburg, Frederik[2]; Eufinger, Jens;[1] Gries, T.[2]

[1] University of Applied Science Darmstadt, Schöfferstraße 3, 64295 Darmstadt, Germany
[2] Institut für Textiltechnik, RWTH Aachen University, Otto-Blumenthal-Straße 1, 52074 Aachen, Germany

**Abstract.** In manufacturing sectors such as textiles and electronics, manual processes are a fundamental part of production. The analysis and monitoring of the processes is necessary for efficient production design. Traditional methods for analyzing manual processes are complex, expensive, and inflexible. Compared to established approaches such as Methods-Time-Measurement (MTM), machine learning (ML) methods promise: Higher flexibility, self-sufficient & permanent use, lower costs. In this work, based on a video stream, the current motion class in a manual assembly process is detected. With information on the current motion, Key-Performance-Indicators (KPIs) can be derived easily. A skeleton-based action recognition approach is taken, as this field recently shows major success in machine vision tasks. For skeleton-based action recognition in manual assembly, no sufficient pre-work could be found. Therefore, a ML pipeline is developed, to enable extensive research on different (pre-) processing methods and neural nets. Suitable well generalizing approaches are found, proving the potential of ML to enhance analyzation of manual processes. Models detect the current motion, performed by an operator in manual assembly, but the results can be transferred to all kinds of manual processes.

**Keywords:** Action Recognition, Manual Assembly, Motion Classification, Skeleton-based, Machine Learning Pipeline

## 1    Introduction

In manufacturing sectors such as textiles and electronics, manual - i.e., human - processes are a fundamental part of production. The complexity, versatility and diversity of some operations do not allow for efficient automation [1]. The increase in individualized products further increases process complexity [2]. Economic efficiency, quality and safety requirements make controls and analyses of all process steps necessary. Key performance indicators (KPIs) are determined, which evaluate processes in terms of various metrics and make them comparable: Time, quality, efficiency and many more [3, 4, 5].

Manual processes are prone to errors and irregularities, and at the same time they are difficult to monitor [1, 6, 7, 8]. A traditional method for analyzing manual processes is Methods Time Measurement (MTM). For this, processes are broken down into small components. For these small, standardized motion sequences, detailed times and formulas for their composition exist [9, 10]. The MTM method maps processes effectively. However, the application is expensive, complex, time-consuming, and inflexible: trained personnel is required, analyses lose their validity even with small process changes [8].

A study by Fraunhofer-Gesellschaft e.V., Munich, shows the potential and areas of application of machine learning in production [11]. The potential of ML methods, which derive information based on human actions, is presented in many works, e.g. [12, 13, 14, 15]. For the analysis of manual processes, methods of the ML field action recognition can be used. Action recognition has been receiving increasing attention and improvements since 2016 [16]. The meta-study [16] shows current developments and trends: Solutions can recognize people and their poses or identify moving areas in videos, and much more. [17, 18, 19, 20, 21].

For the analysis of manual processes, no ML-based method has been established yet. It is assumed that ML-based methods for the analysis of manual processes can be developed based on action recognition approaches. ML methods promise compared to classical approaches: Higher flexibility (e.g., in case of production changes), self-sufficient and permanent use, lower costs and higher precision [22, 23, 24].

## 2   State of technology

The presented work touches two areas of research of which a fundamental understanding is given in the following.

### 2.1   Analyses of manual processes

Manual processes in sectors like textiles and electronics are a key part of the production. To optimally design, understand and improve these processes, they need to be analyzed. KPIs can be used to do so. KPIs can vary from production volume over station utilization to cycle times and many more. [25] The goal of analyzing a process is usually first to determine KPIs. The determination of KPIs in turn aims at optimizing the production with regard to any KPI by means of an acquired and deepened understanding of the process. [4]

Manual processes are often more complex and longer than automated machine steps. The increase in variant versus mass production further increases process complexities and makes manual processes indispensable. The analysis of manual processes is therefore an important part of production design and optimization [6, 7].

As the processes itself are complex, so is the analysis. In contrast to machines, humans do not have standardized interfaces to gain information from, nor are movements always the same. When it comes to changing workers on the same station, comparison difficulties grow, as different humas have different physical

compositions. Moreover, movements, errors and needed times change with the state of fatigue over a shift. [9]

To design and analyze manual processes in the context of production, the Methods-Time-Measurement (MTM) method has become particularly established. Activities are broken down into small sub-movements. For these small, standardized sequences of movements, detailed times and formulas for their composition exist. The tables used for this are called standard time value charts; they were developed in the 1960s through comprehensive studies. Since then, they have been an important part of assembly planning. The disadvantage of the MTM method is its complexity and the need for trained experts. MTM is therefore expensive. Moreover, results are not reusable when a process is modified. A methodology comparable to MTM exists from Verband für Arbeitsgestaltung, Betriebsorganisation und Unternehmensentwicklung e. V. (REFA). [9, 10, 26, 27]

With the increase in digitalization, methods for analyzing manual processes are increasingly appearing, that are no longer based on standard time value maps:

Two patents were filed in 2018 by Amazon.com Inc, Seattle, USA, to guide department stores' workers and monitor their activity based on an ultrasonic wristband [28].

[21] provides a system based on portable location sensors (wearables) that analyses manual processes about cycle times, route optimization, ergonomics and more.

[14] uses multiple wearable sensors (attached to a full-body suit and gloves) to analyze manual processes. Basic movement sequences are detected, comparable to MTM approaches.

A camera-based system for process analysis is offered by [29]. An image section is assigned to a process. By pixel contrast comparison at the borders of the process area, the passage of an object, for example, is detected.

[15] presents a camera-based system that detects a worker in his environment and reproduces him in 3d. Approaches for analyzing the process based on the generated data are shown, to determine cycle times, movement sequences and movement speeds.

### 2.2 Action recognition, skeleton based

ML applications can be divided into different areas. Machine vision describes one part of ML applications. The aim of machine vision is to emulate human perception in vision [30]. Depending on the approach, action recognition can be a sub-area of machine vision. Action recognition deals with the understanding of movements. The data used for action recognition is often images or videos. The goals can be the understanding of movement contexts, the recognition of relevant areas (to which, for example, a human would direct his perception) or the classification of actions [16].

Terms related to action recognition are hardly used uniformly. The following list gives an overview of different areas and what is usually understood by them [16]:
- Pose Estimation: The recognition of poses, often to emphasize the static component of movement or database

- Motion Classification: The recognition of motions, often to emphasize the dynamic component of motion or the database

[16] is a meta study about action recognition and its development over the past years, it is therefore recommended for a brought introduction to the field. The use of skeleton and pose information for action recognition emerged in 2017 and showed clear potential [16, 31].

Skeletons are skeleton-like representations of e.g., human bodies, faces or hands. A skeleton model provides significant points representing e.g., the described parts of a human. The points are given in two (x and y, relative to an images boarders) or three dimensions (with a relative estimation for the z component). Typically, skeleton models also indicate which point is connected to which, see e.g. [32]. The skeleton is the totality of the points and their connections. Fig. 1 shows skeletons derived with the MediaPipe-Hands solution [17].

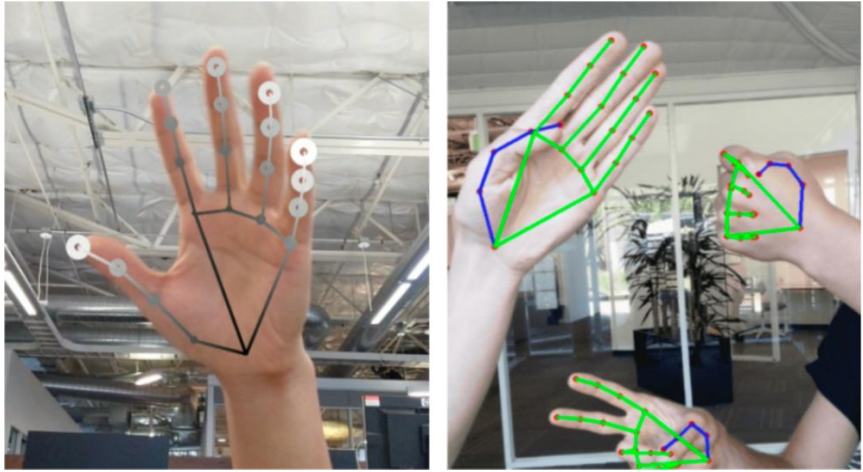

**Fig. 1**. Visualized hand skeletons derived with MediaPipe [17]

In many action recognition problems, the objects represented by the skeletons carry crucial information for the desired goal of an ML model. If this is the case, action recognition models can be developed based on skeleton models. In this case, the skeleton model is the backbone of a multi-stage solution. The skeleton representations per frame are either the only input or an additional input to the original image material, for the next stage, which preforms e.g., motion classification. [16]

Using skeletons vs. images, an input has significant higher information density, which enables:
- Using much smaller models, compared to classical image processing
- Overfitting with a foundation at the high dimensionality of an image does not appear
- Temporal and spatial dependency's can be modeled easily

Simultaneously, good skeleton inputs solve classical generalization hurdles in machine vision based (human) action recognition, like:
- Different lighting situations
- Changing environments
- Differences in the appearance of people: size, clothing, skin color
- Different camera positions

In skeletons-based approaches, an instance can be a sequence of skeleton representations, that corresponds to a sequence of images. The sequence of skeletons is used to classify a motion. The classification problem is then similar to working with time series data, where many machine learning and deep learning architectures are appliable:

[33] successfully uses a simple dense architecture, for classification of gestures. The solution can be used to provide better human-computer interaction.

In [34] the skeletons are transformed into a normalized heatmap. This provides images of normalized skeletons, which can be successfully processed with classical convolutional neural network (CNN) methods. They achieve state-of-the-art performance on multiple benchmarks [35] shows a comparable approach.

[36] and [37] show pure recurrent neural network (RNN) based solutions for action recognition based on skeletons. In [38], RNNs are provided with attention.

[39] and [40] work with a combination of RNN and CNN components, where the CNN components are not used as classical image-processing components but are applied as 1-d convolutions on the representations.

[41] and [42] use graph neural networks (GNNs) or graph convolutional neural networks (GCNs). The applicability of graph-networks is justified by the fact that skeleton representations have similarities to graph structures: Each point has special connections to other points, which are not random but contain ground truth information (i.e., the human body, the anatomy of a hand, etc.).

All the introduced approaches showed competitive results. As an observation, it can be found that some of the architectures are designed very carefully, quite complicated and with a huge number of parameters. For these approaches, often state-of-the-art (SOTA) performance is achieved on a specific problem. On the other hand, some works show simple networks, which do not lead to SOTA performance, but to a better generalization at varying tasks.

Common difficulties faced in action recognition are [16]:
- <u>Determination of motion classes:</u> Currently no norm exists, clarifying in how little pattern a motion shall be subdivided
- <u>Transition between classes:</u> At which exact point in time does a motion class change
- <u>Velocities:</u> The same motion can be performed on highly varying speeds, leading to a scale problem
- <u>Generating data:</u> In contrast to object detection, common labeling pattern do not exist (yet)

## 3 Machine learning approach

For a ML action recognition approach to analyze manual processes, described findings and difficulties from §2 need to be regarded. For manual assembly processes, no (open sourced) established approaches could be found. This leads to high insecurities regarding the best hardware, preprocessing, and neural network configurations. To tackle the huge number of possibilities without using pure guessing or extraordinary computational resources, a ML pipeline is built, which enables combined optimization of hardware-, preprocessing and network approaches.

### 3.1 Database

A manual assembly process is used for the work shown in this paper. The process is executed on a table, above which a camera records the process. Fig. 2 shows the receptive field.

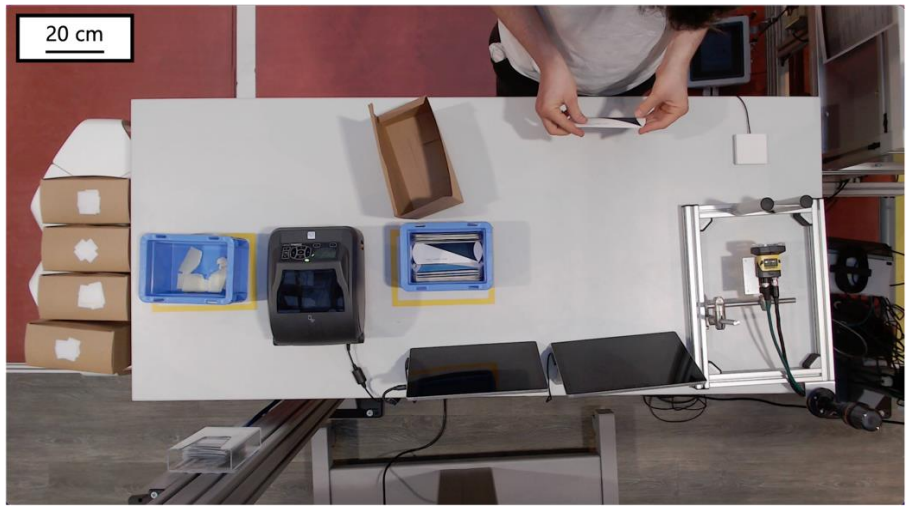

**Fig. 2.** Receptive field for the models

The process scheme is sketched in Fig. 3. The process consists of packing ten times three textile wristbands into shipping boxes. Each's wristband RFID chip is scanned, each band gets quality checked and packed in a single package, before being put into the final shipping's box. The final box is closed, labeled, and handed over to the next process step. The process is split up into ten motion classes (MCs). The nine main MCs and the underlying processes are shown in Fig. 3. MC 0 is an error class, which is used to label any incorrect movements performed unintentionally.

The process is performed by nine people, who need between 13 and 20 minutes for the process. Labeling consists in creating a table, in which the transitions between MCs are defined. Specifically, the frame number of the start of a MC is given. A routine then extracts for every frame a skeleton representation, using the current

SOTA skeleton-hand solution MediaPipe [17]. If a hand is detected, 21 3-dimensional points are saved, along with confidence scores of MediaPipe. If a hand is not detected, all values default to NaNs. That way, for every frame from the original video, a hand representation is saved along with the label, which is the current MC. This way, 240,000 labeled skeleton representations are derived. For implementation details, see [43].

Eight of the nine videos are used for train- and validation data, one video is considered as holdout set. An 80:20 train-validation split is used, which is applied to every video individually after 80 % of its duration. It's worth noting, that MediaPipe detected on average one hand per image, while for the very most images two hands are expected to be visible. A better implementation might result in accuracy gains of the following work but was not part of the presented research.

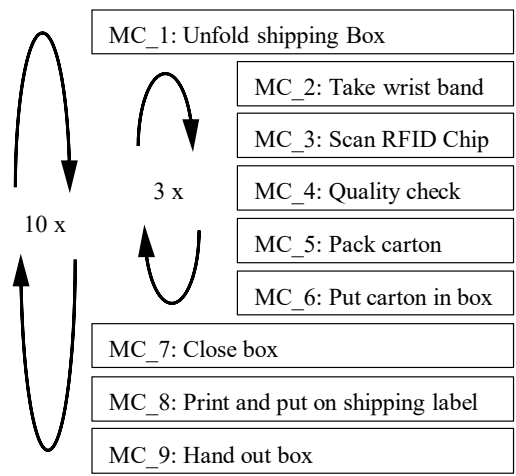

**Fig. 3.** Motion classes (MCs) in the manual example process

### 3.2 Pipeline

To efficiently find promising approaches, a pipeline is built, that contains everything from simulating different fps-rates in the input data, different manipulation and (pre-) processing approaches to building a neural network. The described database serves as input for the pipeline, MCs are the output.

The pipelines concept is shown in Fig. 4. The main block (yellow, marked with an A) receives a set of parameters. The up to 26 hyperparameters (concrete number depends on the architecture) determine e.g., which framerate is emulated, which normalization method is used, and which network architecture is chosen. Based on the parameters, the pipeline builds an end-to-end model. Parameters given can be divided in different category, based on what they control. The categories are shown as yellow clouds, connected to main building blocks of the pipeline where they tune which strategies and methods are used. The pipeline also ensures logging, model saving, cloud synchronization and some more chore functions.

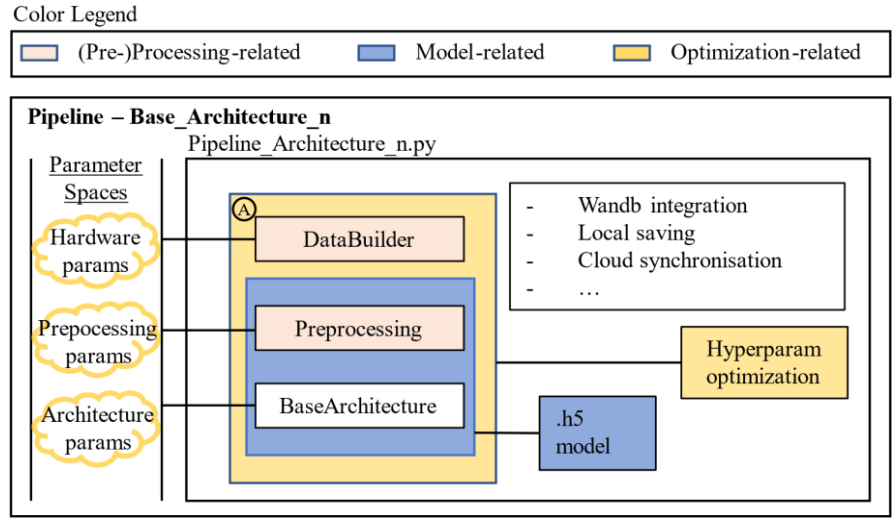

**Fig. 4.** Pipeline concept

'DataBuilder' describes a block, that emulates different camera framerates and the number of pictures used for each instance. The block uses the tf.data API [44] to efficiently prefetch and output the data.

The 'Preprocessing' block contains four custom layers: 1.) Skeleton swapping layer, which ensures hands detected as left and right are always found on the same indices in the instance tensor. For now, it operates based on handedness confidence scores. Location based approaches might be used as alternative or as optimization choice. 2.) Imputing layer, which determines the imputing strategy. At the time of writing, missing data is filled with the constant values. This aims at teaching the model to detect and use the information, that for some movements hands are not detected. Promising methods that are targeted in future work are to predict the skeletons for missing data or to interpolate for elapsed data. This will add one more dimension to the optimization space. 3.) Dimension reduction layer. Based on the parameter set, skeleton dimensions are kept, reduced to the center of gravity, reduced to 5 significant points and more. 4.) Normalizing layer. Based on the parameter set, skeletons are normalized a) between 0 and 1 relative to the image borders, b) each instance skeletons are normalized on the most current skeleton or c), every skeleton is normalized on itself. The grade of abstraction/generalization increases from a) to c)

The 'BaseArchitecture' block builds one of three basic neural networks, suitable for processing time series data: 1.) Long-short-term-memory (LSTM), with up to 20 layers of which each contains up to 250 cells. 2.) Time-distributed dense (TD-dense) with up to 20 flat- and 20 time-distributed-layers, of which each contains up to 500 cells. The optimization is also allowed to neglect TD or flat layers completely. 3.)

One dimensional convolution (Conv-1d) with up to 10 layers. The layers may all have the same number of filters with up to 128 or double the number of filters every layer. Strides can be 1 to 5 with padding causal or same. Moreover, the optimization is allowed to activate pooling at the end into 1 to 120 sections per filter.

The basic architectures are initialized with their different alterable variables, are initialized depending on the given parameter set. It is to be noted, that graph neural networks also seem quite appropriate, see e.g., [41, 42]. As no standardized keras implementations are released yet, they were not included due to time constrains.

The pipelines result is a trained TensorFlow .h5 model and a parameter file. The 'DataBuilder' block is not included in the .h5 model, as its parameters found, determine how the hardware is set up. In the projects public repo, methods are included to immediately use an achieved model to predict new data [43]. The pipeline is built using TensorFlow.

The goal was to not only optimize through a fixed set of options but to create a pipeline, that is easily extendable. We state, that with little to no effort, any additional method can be implemented. Design choice and (pre-) selection of methods are based on the current state of technology, e.g. [22, 23, 30].

## 4   Optimization

The Parameters given to the pipeline are treated as hyperparameters. Bayesian optimization is used to vary the parameters. Bayesian optimization allows to optimize through the space of 26 parameters, considered as hyperparameters. To speed up convergence, multiple optimization runs are done while:
- Spaces for parameters are inclemently reduced
- Converged parameters are frozen
- The selectable learning rates are decreased
- Max epoch number is increased

Bayesian optimization is implemented via weights and biases (wandb) [45]. The tools of wandb are used to identify promising parameters spaces and converged parameters (sensitivity analyses, parallel coordinate plots, comparisons of run metrics).

For each base architecture (LSTM, TD-dense, Conv-1d) optimization is performed individually. That way, the dimensions of the hyperparameter space are further reduced.

## 5   Results

The developed pipeline allows to optimize method combinations and find models that fit desired needs inference prediction time, model generalization and more. This is achieved by restricting desired methods and paths in the pipeline during optimization to only allow finding models, that fit individual needs.

The normalization methods for example vary from keeping the absolute coordinates of the hands (models can now use the information of the position of the hand in the images) to normalize each hand on itself (models now only know how a hand is aligned to itself for each frame). The later method offers bigger generalization, as a model does not link motions to image sections and thus is not sensitive if a process is performed in another region of the camera. On the other hand, neglecting location information complicates the model's learning.

As it became clear in the early optimization phase, that also methods with high generalization potential can yield competitive results (accuracies > 75 %), we choose to limit the optimization to methods with high generalization potential. The location information was concealed, and models were not allowed to use data from more than 3,5 seconds in the past. If a model only relies on more current data, it swings back in quicker if a mistake is performed by the worker and learns less information about which motions follow each other, what again increases generalization.

2,158 trainings were performed, which took 6 weeks using a RTX 3080Ti. All separate optimizations for the three base architectures converged. Best models reached validation accuracies around 80 %.

Best results were achieved using the maximum available camera framerate of 30 fps, which was expected, as more pictures contain more information. Reducing the amount of coordinate representations slightly diminish model performance. Looking at models with validation accuracies > 76 %, model sizes vary from 19,000 to 10,000,000 parameters. Specific well-working-together parameter combinations are shown below for a specific model.

In further works, routines will be developed, that use the motion class predictions to extract KPIs like the cycle time in-situ. To suggest a specific model for future works, models received from the optimization are compared. As main criteria, good and consistent performance over different workers is requested. We decided to compare models from all base architectures (with the highest grade of generalization) along with two models that use less generalizing methods (their skeleton inputs contain location information). A model is pre-selected if it has a per-optimization top validation accuracy and differs from other already pre-selected models of the same base architecture. This way, a pre-selection of 11 models is derived. Fig. 5 shows the models and their accuracies over the full train- and validation-set as well as over the specific workers that form the datasets.

It is to be noted, that worker (w.) 4, 7 and 8 operated with many inconsistencies compared to the others, whereas w. 3, 4 and 5 operated the cleanest. Moreover, data from w.-9 was not used during development and thus gives a hint on the models performance on workers it was not trained on.

We observe, that 1), the specialized models yield the best performance on sloppy data. Which is unsurprisingly, as the spatial information can be learned and used. 2), The LSTM models fail to generalize well on workers they have not seen in training. 3) The models performance is significantly boosted when motions are performed consistently, which can be assumed when used in production. 4) The TD-Dense models yield the best and most consistent performance across different workers (excluding the specialized models). While using the highest generalization methods implemented in the pipeline, TD-Dense model 1 and 2 perform only 3 % worse compared to specialized models on the holdout data.

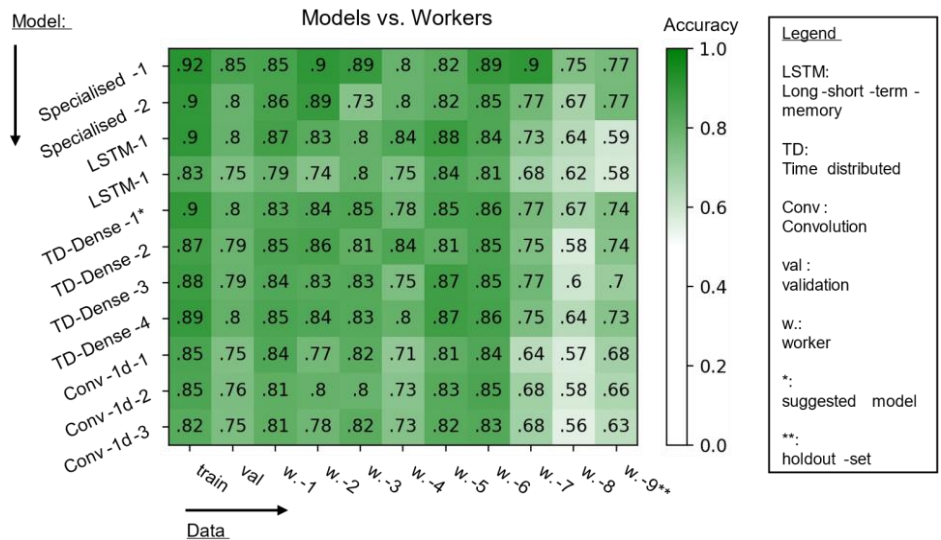

**Fig. 5.** Comparing the performance of models with different base architectures on different workers

TD-Dense-1 model is suggested as final model. It is considered the strongest Dense-TD model with best performance on the holdout and unclean datasets. We value the generalization over the accuracy gain if choosing a specialized model. TD-Dense-1 has a validation accuracy 80,4 %, holdout accuracy 74 %. The model consists of 11 time-distributed dense layers with 188 cells and 2 flat dense layers with 457 cells. Instances consist of skeletons from the past 104 images that are recorded @30-fps. Each skeleton is normalized on itself, missing data is imputed with the constant 2. Training was started with a learning rate of 10e-4 that got reduced on plateaus. The model was trained for 32 epochs, using categorical cross-entropy loss and the Adam optimizer with keras defaults.

Fig. 6 shows the per-class f-scores and support for the final model. Multiple insights can be derived:

1) Using an error class (class 0) did not lead to the desired outcome and is taken as fail. While having a precision of 65 %, it's recall is below 10 %. We suggest implementing an anomaly detection model in inference. 2) Higher support yields better scores. Therefore, we argue, with more training data for underrepresented classes (e.g., 2, 6, 7, 9) higher accuracies might be possible. 3) Indeed, the previous mentioned is not true for class 8, which achieves a better score than class 5, although it relies on less than one third of the data. This may mean, some motions are more significant and therefore easier to predict. 4) As the accuracy is imbalanced between classes, some yield significant higher accuracies than the overall accuracy. Class 5 reaches the highest accuracy with 93 %. For developing routines that extract KPIs like the cycle time based on the motion classification, for the given process and the given model, this class is a solid starting point.

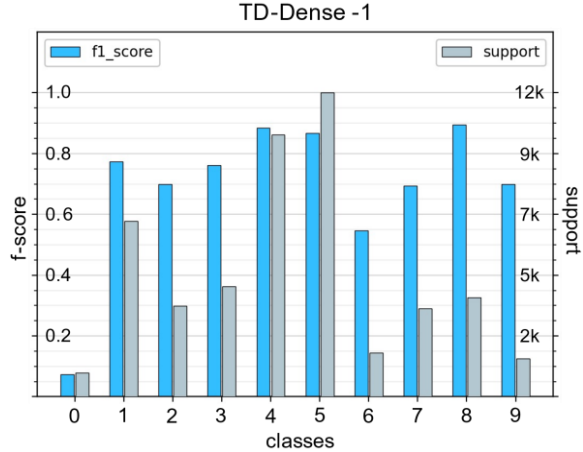

**Fig. 6.** Per-class f-scores and support for the final model on the validation set

With routines presented in [43], we further investigated the average accuracy per class over the time of an action appearance. The average accuracy reaches its maximum at the middle-to-end section of an action appearance, which is a valuable inside for the KPI routines. Furthermore, we investigated the errors made. For most occasions, confusions happen in the transition phase between motions and classes confused are the consecutive ones. We argue this shows the model achieved a general understanding of the motions. Moreover, we experimented with different persons labeling the transition between two motions and determined deviations of up to one second. It is hard to tell whether the model makes mistakes here or if the labeling is noisy. To tackle this issue, one could iteratively train models and use them for setting the transitions between motion classes.

While visualizing the predictions over the video footage, one can obtain the advantage of using sequential input data and not only one image: For some parts, multiple frames do not yield skeleton detections. Thanks to the sequential inputs, the model is still able to make solid predictions for each frame.

## 6   Conclusion and discussion

We presented a skeleton-based ML approach, to detect motions in manual processes. As main result we state, skeleton-based action recognition can be used to automate the analysis of manual processes. We state, the results of our skeleton-based action recognition approach are good, which encourages us, to further develop routines, that use the predicted motion classes to derive production KPIs. We showed that not one specific method, but the right combination of different methods leads to strong ML models. Therefore, we value the developed pipeline as much as the specific final TD-Dense model. The final model uses the highest generalization possible with the current pipeline and achieves a validation accuracy of 80.4 % and a holdout accuracy of 74 %. For a prediction, the video material of the past 3.5 seconds @30-fps is used.

Each instance contains the skeleton data of these images, extracted using the MediaPipe open-source solution.

We expect that without adaption, our pipeline can be applied to any manual process, if the workers hands make up the significant part of the motions. Moreover, the pipeline is easily extendable to integrate any (pre-) processing method. Unexplored but promising methods are: trained imputing strategies, graph neural networks or the combination of ML elements that are currently separated into different base architectures. Moreover, we assume to receive stronger models by applying common ML techniques like model ensembles, integrating an unsupervised backbone model that is capable of understanding hand motions, improve the skeleton predictions or add a classic convolutional image processing data path.

The usability of the presented solution highly depends on the success of the KPI extracting routines, that are to be developed on top of the presented model. The main difficulty to tackle there is to reliably get the currently active motion class, as the ground truth is continuous, whereas the predictions are discrete and scattered, especially around motion class transitions. We expect this problem to be solvable.

ML is increasingly used in many branches to automate tasks. The solutions typically have lower costs and higher accuracies. The analysis of manual processes is a branch where ML is not established yet. Our work proves the potential of skeleton-based action recognition to be the ML technique, with which manual processes automatically can be analyzed. Our pipeline can be used to easily get a ML model for any manual process.